\documentclass[runningheads]{llncs}
\usepackage{graphicx}
\usepackage{amsmath,amssymb} 
\usepackage{color}
\usepackage{algorithm}
\usepackage{algorithmic}
\usepackage[pdftex, pagebackref, colorlinks, citecolor=blue, linkcolor=blue]{hyperref}
\usepackage{booktabs}

\begin{document}

\title{Unseen Object Segmentation in Videos \\via Transferable Representations} 
\titlerunning{Unseen Object Segmentation in Videos via Transferable Representations} 


\author{Yi-Wen Chen$^{1,2}$,
Yi-Hsuan Tsai$^{3}$,
Chu-Ya Yang$^{1}$,\\
Yen-Yu Lin$^{1}$,
Ming-Hsuan Yang$^{4,5}$}

%

\authorrunning{Y.-W. Chen et al.} 


\institute{$^{1}$Academia Sinica,
$^{2}$National Taiwan University,
$^{3}$NEC Laboratories America,
$^{4}$University of California, Merced,
$^{5}$Google Cloud}

\maketitle

\begin{abstract}
In order to learn object segmentation models in videos, conventional methods require a large amount of pixel-wise ground truth annotations.
However, collecting such supervised data is time-consuming and labor-intensive.
In this paper, we exploit existing annotations in source images and transfer such visual information to segment videos with unseen object categories.
Without using any annotations in the target video, we propose a method to jointly mine useful segments and learn feature representations that better adapt to the target frames.
The entire process is decomposed into two tasks: 1) solving a submodular function for selecting object-like segments, and 2) learning a CNN model with a transferable module for adapting seen categories in the source domain to the unseen target video.
We present an iterative update scheme between two tasks to self-learn the final solution for object segmentation.
Experimental results on numerous benchmark datasets show that the proposed method performs favorably against the state-of-the-art algorithms.
%
\end{abstract}
\section{Introduction}
Nowadays, video data can be easily accessed and visual analytics has become an important task in computer vision.
In this line of research, video object segmentation is one of the effective approaches to understand visual contents that 
facilitates various applications, such as video editing, content retrieval, and object identification.
While conventional methods rely on the supervised learning strategy to effectively localize and segment objects in videos,
collecting such ground truth annotations is expensive and cannot scale well to a large amount of videos.

Recently, weakly-supervised methods for video object segmentation \cite{Tsai_ECCV_2016,Zhang_CVPR_2017,Saleh_ICCV_2017,Yan_CVPR_2017} have been developed to relax the need for annotations where only class-level labels are required.
These approaches have significantly reduced the labor-intensive step of collecting pixel-wise training data on target categories.
However, these categories are pre-defined and thus the trained model cannot be directly applied to unseen categories in other videos,
and annotating additional categories would require more efforts, which is not scalable in practice.
In this paper, we propose an algorithm to reduce efforts in both annotating pixel-level and class-level ground truths 
for unseen categories in videos.

To this end, we make use of existing pixel-level annotations in images from the PASCAL VOC dataset~\cite{PASCAL_VOC_2010} with pre-defined categories, and design a framework to transfer this knowledge to unseen videos.
That is, the proposed method is able to learn useful representations for segmentation from the data in the image domain and adapt these representations to segment objects in videos regardless of whether their categories are covered in the PASCAL VOC dataset.
Thus, while performing video object segmentation, our algorithm does not require annotations in any forms, such as pixel-level or class-level ground truths.

We formulate the object segmentation problem for unseen categories as a joint objective of mining useful segments from videos while learning transferable knowledge from image representations.
Since annotations are not provided in videos, we design an energy function to discover object-like segments in videos based on the feature representations learned from the image data.
We then utilize these discovered segments to refine feature representations in a convolutional neural network (CNN) model, while a transferable module is developed to learn the relationships between multiple seen categories in images and the unseen category in videos.
By jointly considering both energy functions for mining better segments while learning transferable representations, we develop an iterative optimization method to self-guide the video object segmentation process.
We also note that the proposed framework is flexible as we can input either weakly-labeled or unlabeled videos.

To validate the proposed method, we conduct experiments on benchmark datasets for video object segmentation.
First, we evaluate our method on the DAVIS 2016 dataset~\cite{DAVIS2016}, where the object categories may be different from the seen categories on PASCAL VOC.
Based on this setting, we compare with the state-of-the-art methods for object segmentation via transfer learning, including approaches that use the NLP-based GloVe embeddings~\cite{Pennington_EMNLP_2014} and a decoupled network \cite{Hong_CVPR_2016}.
In addition, we show baseline results with and without the proposed iterative self-learning strategy to demonstrate its effectiveness.
Second, we adopt the weakly-supervised setting on the YouTube-Objects dataset~\cite{Prest_CVPR_2012} and show that the proposed method performs favorably against the state-of-the-art algorithms in terms of visual quality and accuracy.

The contributions of this work are summarized as follows.
First, we propose a framework for object segmentation in unlabeled videos through a self-supervised learning method.
Second, we develop a joint formulation to mine useful segments while adapting the feature representations to the target videos.
Third, we design a CNN module that can transfer knowledge from multiple seen categories in images to the unseen category in videos.
%
%
\section{Related Work}
{\flushleft {\bf Video Object Segmentation.}}
Video object segmentation aims to separate foreground objects from the background.
Conventional methods utilize object proposals \cite{Lee_ICCV_2011,Perazzi_CVPR_2015,Koh_CVPR_2017} or graphical models \cite{Tsai_CVPR_2016,Marki_CVPR_2016}, while recent approaches focus on learning CNN models from image sequences with frame-by-frame pixel-level ground truth annotations to achieve the state-of-the-art performance \cite{Cheng_ICCV_2017,Tokmakov_ICCV_2017,Jain_CVPR_2017}.
For CNN-based methods, motion cues are usually used to effectively localize objects.
Jain et al.~\cite{Jain_CVPR_2017} utilize a two-stream network by jointly considering appearance and motion information.
The SegFlow method~\cite{Cheng_ICCV_2017} further shows that jointly learning segmentation and optical flow in videos enhances both performance.
Another line of research is to fine-tune the model based on the object mask in the first frame \cite{Caelles_CVPR_2017,Khoreva_cvpr_2017} and significantly improves the segmentation quality.
However, in addition to annotations of the first frame in target videos \cite{Caelles_CVPR_2017,Khoreva_cvpr_2017}, these methods require pre-training on videos with frame-by-frame pixel-level annotations  \cite{Cheng_ICCV_2017,Tokmakov_ICCV_2017} or bounding box ground truths \cite{Jain_CVPR_2017} to obtain better foreground segmentation.
In contrast, the proposed algorithm uses only a smaller number of existing annotations from the image dataset and transfers the feature representations to unlabeled videos for object segmentation.
In addition, our method is flexible for the weakly-supervised learning setting, which cannot be achieved by the above approaches.
%
{\flushleft {\bf Object Segmentation in Weakly-supervised Videos.}}
To reduce the need of pixel-level annotations, weakly-supervised methods have been developed to facilitate the segmentation process, where only class-level labels are required in videos.
Numerous approaches are proposed to collect useful semantic segments by training segment-based classifiers \cite{Tang_CVPR_2013} or ranking supervoxels \cite{Zhong_ACCV_2016}.
However, these methods rely on the quality of generated segment proposals and may produce inaccurate results when taking low-quality segments as the input.
Zhang et al.~\cite{Zhang_CVPR_2015} propose to utilize object detectors integrated with object proposals to refine segmentations in videos.
Furthermore, Tsai et al.~\cite{Tsai_ECCV_2016} develop a co-segmentation framework by linking object tracklets from all the videos and improve the result.
Recently, the SPFTN method~\cite{Zhang_CVPR_2017} utilizes a self-paced learning scheme to fine-tune segmentation results from object proposals.
Different from the above algorithms that only target on a pre-defined set of categories, our approach further extends this setting to videos without any labels for unseen object categories.
%
{\flushleft {\bf Transfer Learning for Object Recognition.}}
Using cross-domain data for unsupervised learning has been explored in domain adaptation \cite{Saenko_ECCV_2010,Gopalan_ICCV_2011,Patricia_CVPR_2014,Ganin_ICML_2015}.
While domain adaptation methods make the assumption that the same categories are shared across different domains,
transfer learning approaches focus on transferring knowledge between categories.
Numerous transfer learning methods have been developed for object classification \cite{Tommasi_PAMI_2014} and detection \cite{Lim_NIPS_2011,Hoffman_NIPS_2014}.
Similar efforts have been made for object segmentation.
Hong et al.~\cite{Hong_CVPR_2016} propose a weakly-supervised semantic segmentation method by exploiting pixel-level annotations from different categories.
Recently, Hu et al.~\cite{Hu_arxiv_2017} design a weighted transform function to transfer knowledge between the detected bounding boxes and instance segments.
In this work, we share the similar motivation with \cite{Hong_CVPR_2016} but remove the assumption of weak supervisions.
To the best of our knowledge, this work is the first attempt for video object segmentation by transferring knowledge from annotated images to unlabeled videos between unshared categories.
\begin{figure}[t]
	\centering
	\includegraphics[width=0.85\linewidth]{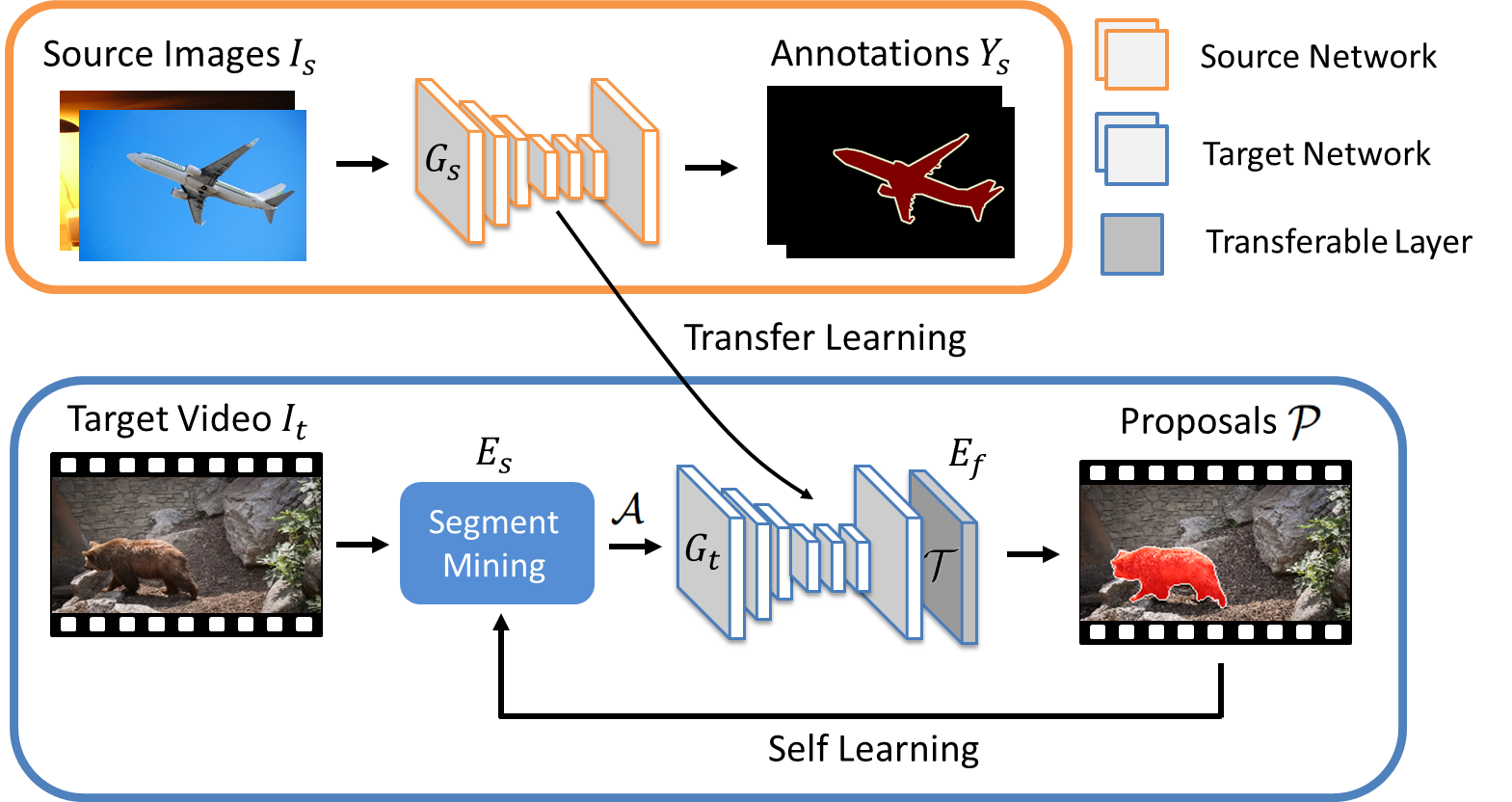}
	\caption{Overview of the proposed algorithm. Given a set of source images $\mathcal{I}_s$ with semantic segmentation annotations $Y_s$, we first train a source CNN model $G_s$.
		To predict object segmentations on a target video $\mathcal{I}_t$ without knowing any annotations, we initialize the target network $G_t$ from the parameters in $G_s$ and perform adaptation via a transferable layer $\mathcal{T}$.
		We optimize the function $E_s$ for selecting object-like segments $\mathcal{A}$ from proposals $\mathcal{P}$ and adapt feature representations in the CNN model via optimizing $E_f$.
		The entire self-learning process is performed via iteratively updating two energy functions to obtain the final segmentation results.
	}
	\label{fig:overview}
\end{figure}
\section{Algorithmic Overview}
%
%
\subsection{Overview of the Proposed Framework}
We first describe the problem context of this work.
Given a number of source images $\mathcal{I}_s = \{I_s^1, ..., I_s^N\}$ with pixel-level semantic segmentation annotations $Y_s = \{y_s^1, ..., y_s^N\}$ and the target sequence $\mathcal{I}_t = \{I_t^1, ..., I_t^M\}$ without any labels, our objective is to develop a self-supervised learning algorithm that segments the object in $\mathcal{I}_t$ by transferring knowledge from $\mathcal{I}_s$ to $\mathcal{I}_t$.
In this work, the object category in $\mathcal{I}_t$ is allowed to be arbitrary. It can be either covered by or different from those in $\mathcal{I}_s$.
%

To this end, we propose a method with two components: 1) a ranking module for mining segment proposals, and 2) a CNN model for learning transferable feature representations.
Fig.~\ref{fig:overview} illustrates these two components in the proposed framework.
We first train a source CNN model $G_s$ using $\mathcal{I}_s$ and $Y_s$ as the input and the desired output, respectively.
Then we initialize the target network $G_t$ from the parameters in $G_s$, where this target network can generate segment proposals $\mathcal{P}$ on the target video $\mathcal{I}_t$.
To find a set of object-like proposals among $\mathcal{P}$, we then develop an energy function to re-rank these proposals based on their objectness scores and mutual relationships.
With the selected proposals that have higher object-like confidence, we further refine the feature representations in the target network.
Since $\mathcal{I}_s$ and $\mathcal{I}_t$ may not share common object categories, we design a layer $\mathcal{T}$ that enables cross-category knowledge transfer, and append it to the target network.
The entire process can be formulated as a joint optimization problem with the objective function as described below.
\subsection{Objective Function}
Our goal is to find high-quality segment proposals $\mathcal{P}$ from the target video $\mathcal{I}_t$ that can guide the network to learn feature representations $\mathcal{F}$ for better segmenting the given video $\mathcal{I}_t$.
We carry out this task by jointly optimizing an energy function $E$ that accounts for segment proposals $\mathcal{P}$ and features $\mathcal{F}$:
\begin{equation}
\underset{\mathcal{A}, \theta}{\max} \:  E(\mathcal{I}_t, \mathcal{P}, \mathcal{F}; \mathcal{A}, \theta) =
\underset{\mathcal{A}, \theta}{\max} \: E_s(\mathcal{P}, \mathcal{F}; \mathcal{A}) + E_f(\mathcal{I}_t, \mathcal{A}; \theta),
\label{eq:joint}
\end{equation}
where $E_s$ is the energy for selecting a set of high-quality segments $\mathcal{A}$ from the proposals $\mathcal{P}$ based on the features $\mathcal{F}$, while $\theta$ is the parameters of the CNN model that aims to optimize $E_f$ and learn feature representations $\mathcal{F}$ from the selected proposals $\mathcal{A}$.
Details of each energy function and the optimization process are described in the following section.
\section{Transferring Visual Information for Segmentation}
In this section, we describe the details of the proposed energy functions for mining segments and learning feature representations, respectively.
The segment mining step is formulated as a submodular optimization problem, while the feature learning process
is completed through a CNN with a transferable module.
After introducing both energy functions, we present an iterative optimization scheme to jointly maximize the objective~\eqref{eq:joint}.
\subsection{Mining Segment Proposals}
\label{sec:sub}
Given a target video $\mathcal{I}_t$, we can generate frame-by-frame object segmentations by applying the CNN model pre-trained on the source images $\mathcal{I}_s$.
However, these segments may contain many false positives that do not well cover objects.
Thus, we aim to select high-quality segments and eliminate noisy ones from the generated object segmentations.
The challenging part lies in that there are no ground truth annotations in the target video, and thus we cannot train a classifier to guide the selection process.

Motivated by the co-segmentation method~\cite{Tsai_ECCV_2016}, we observe that high-quality segments have higher mutual relationships.
As a result, we gather all the predicted segments from the target video and construct a graph to link each segment.
We then formulate segment mining as a submodular optimization problem, aiming to select a subset of object-like segments that share higher similarities.
{\flushleft {\bf Graph Construction on Segments.}}
We first feed the target video $\mathcal{I}_t$ into the CNN model frame-by-frame and obtain a set of segment proposals $\mathcal{P}$, where each proposal is a connected-component in the predicted segmentation.
Then we construct a graph $G = (\mathcal{V}, \mathcal{E})$ on the set $\mathcal{P}$, where each vertex $v \in \mathcal{V}$ is a segment, and each edge $e \in \mathcal{E}$ models the pairwise relationship between two segments.
Our goal is to find a subset $\mathcal{A}$ within $\mathcal{P}$ that contains proposals with higher object-like confidence.
{\flushleft {\bf Submodular Function.}}
We design a submodular function to find segments that meet the following criteria: 1) objects from the same category share similar features, 2) a true object has a higher response from the output of the CNN model, and 3) an object usually moves differently from the background area in the video.
Therefore, we formulate the objective function for selecting object-like segments by a facility location term $\mathcal{H}$ \cite{Lazic_ICCV_2009} and a unary term $\mathcal{U}$.
The former computes the similarity between the selected segments, while the latter estimates the probability of each selected segment being a true object. Both terms are defined based on the segment proposals $\mathcal{P}$ and the adopted feature representation $\mathcal{F}$.
First, we define the facility location term as:
\begin{equation}
\mathcal{H}(\mathcal{P}, \mathcal{F}; \mathcal{A}) = \sum_{i \in \mathcal{A}} \sum_{j \in \mathcal{V}} \: W(v_i, v_j)
- \sum_{i \in \mathcal{A}} \phi_i,
\label{eq:sub_facility}
\end{equation}
where $W$ denotes the pairwise relationship between a potential facility $v_i$ and a vertex $v_j$, while $\phi_i$ is the cost to open a facility, which is fixed to a constant $\alpha$.
We define $W$ as the similarity between two segments in order to encourage the submodular function to choose a facility $v_i$ that is similar to $v_j$.
To estimate this similarity, we represent each segment as a feature vector and compute their inner product of the two vectors.
%
To form the feature vector for each segment, we draw feature maps from the CNN model (\textbf{conv1} to \textbf{conv5}) and perform the global average pooling on each segment.
It is the adopted feature representation $\mathcal{F}$ in this work.
In addition to the facility location term, we employ a unary term to evaluate the quality of segments:
\begin{equation}
\mathcal{U}(\mathcal{P}, \mathcal{F}; \mathcal{A}) = \lambda_o \sum_{i \in \mathcal{A}} \Phi_o(i)
+ \lambda_m \sum_{i \in \mathcal{A}} \Phi_m(i),
\label{eq:sub_unary}
\end{equation}
where $\Phi_o(i)$ is the objectness score that measures the probability of segment $i$ being a true object, and $\Phi_m(i)$ is the motion score that estimates the motion difference between segment $i$ and the background region. $\lambda_o$ and $\lambda_m$ are the weights for the two terms, respectively. 

The objectness score $\Phi_o(i)$ is calculated by averaging the probability map of the CNN output layer on all the pixels within the segment.
%
For the motion score $\Phi_m(i)$, we first compute the optical flow~\cite{Liu_thesis_2009} for two consecutive frames, and then we utilize the minimum barrier distance \cite{Strand_CVIU_2013,Zhang_ICCV_2015} to convert the optical flow into a saliency map, where the larger distance represents a larger motion difference with respect to the background region.
{\flushleft {\bf Formulation for Segment Mining.}}
Our goal is to find a subset $\mathcal{A}$ within $\mathcal{P}$ containing segments that are similar to each other and have higher object-like confidence.
Therefore, we combine the facility location term $\mathcal{H}$ and the unary term $\mathcal{U}$ as the energy $E_s$ in~\eqref{eq:joint}:
\begin{equation}
E_s(\mathcal{P}, \mathcal{F}; \mathcal{A}) = \mathcal{H}(\mathcal{P}, \mathcal{F}; \mathcal{A}) + \mathcal{U}(\mathcal{P}, \mathcal{F}; \mathcal{A}).
\label{eq:sub_combine}
\end{equation}
%
%
We also note that the linear combination of two non-negative terms preserves the submodularity \cite{Zhu_CVPR_2014}.
\subsection{Learning Transferable Feature Representations}
\label{sec:cnn}
Given the selected set of object-like segment proposals, the ensuing task is to learn better feature representations based on these segments.
To this end, we propose to use a CNN model fine-tuned on these segments via a self-learning scheme.
However, since our target video may have a different set of object categories from those in the source domain, we further develop a transfer learning method where a transferable layer is augmented to the CNN model.
With the proposed layer, our network is able to transfer knowledge from seen categories to the unseen category, without the need of any supervision in the target video.

Inspired by the observation that an unseen object category can be represented by a series of seen objects \cite{Rochan_CVPR_2015}, we develop a transferable layer that approximates an unseen category as a linear combination of seen ones in terms of the output feature maps.
In the following, we first present our CNN objective for learning the feature representations based on the selected segment proposals.
Then we introduce the details of the proposed layer for transferring knowledge from the source domain to the target one.
{\flushleft {\bf Objective Function.}}
Given the target video $\mathcal{I}_t$ and selected segment proposals $\mathcal{A}$ as described in Section~\ref{sec:sub}, we use $\mathcal{A}$ as our pseudo ground truths and optimize the target network $G_t$ with parameters $\theta_g$ to obtain better feature representations that match the target video.
Specifically, we define the energy function $E_f$ in~\eqref{eq:joint} as:
\begin{equation}
E_f(\mathcal{I}_t, \mathcal{A};\theta_g, \theta_\mathcal{T}) = - \mathcal{L} ( \mathcal{T}(G_t(\mathcal{I}_t)), \mathcal{A} ),
\label{eq:cnn}
\end{equation}
where $\theta_\mathcal{T}$ is the parameters of the transferable layer $\mathcal{T}$ and $\mathcal{L}$ is the cross-entropy function to measure the loss between the network prediction $\mathcal{T}(G_t(\mathcal{I}_t))$ and the pseudo ground truth $\mathcal{A}$.
We also note that, we use the minus sign for the loss function $\mathcal{L}$ to match the maximization formulation in~\eqref{eq:joint}.
{\flushleft {\bf Learning Transferable Knowledge.}}
Suppose there are $C_s$ categories in the source domain, we aim to transfer a source network $G_s$ pre-trained on the source images $\mathcal{I}_s$ to the target video.
To achieve this, we first initialize the target network $G_t$ using the parameters in $G_s$.
Given the target video $\mathcal{I}_t$, we can generate frame-wise feature maps $R = G_t(\mathcal{I}_t) = \{r_c\}_{c=1}^{C_s}$ through the network with $C_s$ channels, where $r_c$ is the output map of source category $c$.
Since the target category is unknown, we then approximate the desired output map, $r$, for the unseen category as a linear combination of these seen categories through the proposed transferable layer $\mathcal{T}$:
\begin{equation}
r = \mathcal{T}(R) = \sum_{c=1}^{C_s} w_c \: r_c,
\label{eq:transfer}
\end{equation}
where $w_c$ is the weight of the seen category $c$.
Specifically, the proposed transferable layer $\mathcal{T}$ can be performed via a $1 \times 1$ convolutional layer with $C_s$ channels, in which the parameter of channel $c$ in $\theta_\mathcal{T}$ corresponds to $w_c$.

Since $w_c$ is not supervised by any annotations from the target video, the initialization of $w_c$ is critical for obtaining a better combination of feature maps from the seen categories.
Thus, we initialize $w_c$ by calculating the similarity between each source category $c$ and the target video.
For each image in the source and target domains, we extract its feature maps from the \textbf{fc7} layer of the network and compute a 4096-dimensional feature vector on the predicted segment via global average pooling.
By representing each image as a feature vector, we measure the similarity score between source and target images by their
inner product.
%
Finally, the initialized weight $w_c^{init}$ for the category $c$ can be obtained by averaging largest scores on each target frame with respect to the source images:
%
\begin{equation}
w_c^{init} = \frac{1}{| \mathcal{I}_t |} \sum_{i=1}^{| \mathcal{I}_t |} \underset{j}{\max} \: \langle \mathcal{F}_t^i, \mathcal{F}_{s,c}^j \rangle,
\label{eq:sim}
\end{equation}
where $| \mathcal{I}_t |$ is the number of frames in the target video, $\mathcal{F}_t^i \in \mathbb{R}^{4096}$ is the feature vector of the $i$th frame of $\mathcal{I}_t$, and $\mathcal{F}_{s,c}^j \in \mathbb{R}^{4096}$ is the feature vector of the $j$th image of source category $c$.
\subsection{Joint Formulation and Model Training}
Based on the formulations to mine segments \eqref{eq:sub_combine} and learn feature representations \eqref{eq:cnn}, 
we jointly solve the two objectives, i.e., $E_s$ and $E_f$, in~\eqref{eq:joint} by:
\begin{align}
\underset{\mathcal{A}, \theta} {\max} \: E(\mathcal{I}_t, \mathcal{P}, \mathcal{F}; \mathcal{A}, \theta)
& = \underset{\mathcal{A}, \theta}{\max} \: E_s(\mathcal{P}, \mathcal{F}; \mathcal{A}) + E_f(\mathcal{I}_t, \mathcal{A}; \theta) \notag\\
& = \underset{\mathcal{A} , \theta_g, \theta_\mathcal{T}}{\max} \: [\mathcal{H}(\mathcal{P}, \mathcal{F}; \mathcal{A}) + \mathcal{U}(\mathcal{P}, \mathcal{F}; \mathcal{A})]
- \mathcal{L} ( \mathcal{T}(G_t(\mathcal{I}_t)), \mathcal{A} ). \notag\\
\label{eq:final}
\end{align}
%
%
%
%
To optimize~\eqref{eq:final}, we decompose the process into two sub-problems: 1) solving a submodular function for segment mining to generate $\mathcal{A}$, and 2) training a CNN model that optimizes $\theta_g$ and $\theta_\mathcal{T}$ for learning transferable representations.
We adopt an iterative procedure to alternately optimize the two sub-problems. The initialization strategy and the optimization of the two sub-problems are described below.
{\flushleft {\bf Initialization.}}
We first pre-train a source network $G_s$ on the PASCAL VOC training set \cite{PASCAL_VOC_2010} with 20 object categories.
We then initialize the target network $G_t$ from parameters in $G_s$ and the transferable layer $\mathcal{T}$ as described in Section~\ref{sec:cnn}.
To obtain an initial set of segment proposals, we forward the target video $\mathcal{I}_t$ to the target model $G_t$ with $\mathcal{T}$ and generate segments $\mathcal{P}$ with their features $\mathcal{F}$.
\begin{figure}[t]
	\centering
	\includegraphics[width=0.9\linewidth]{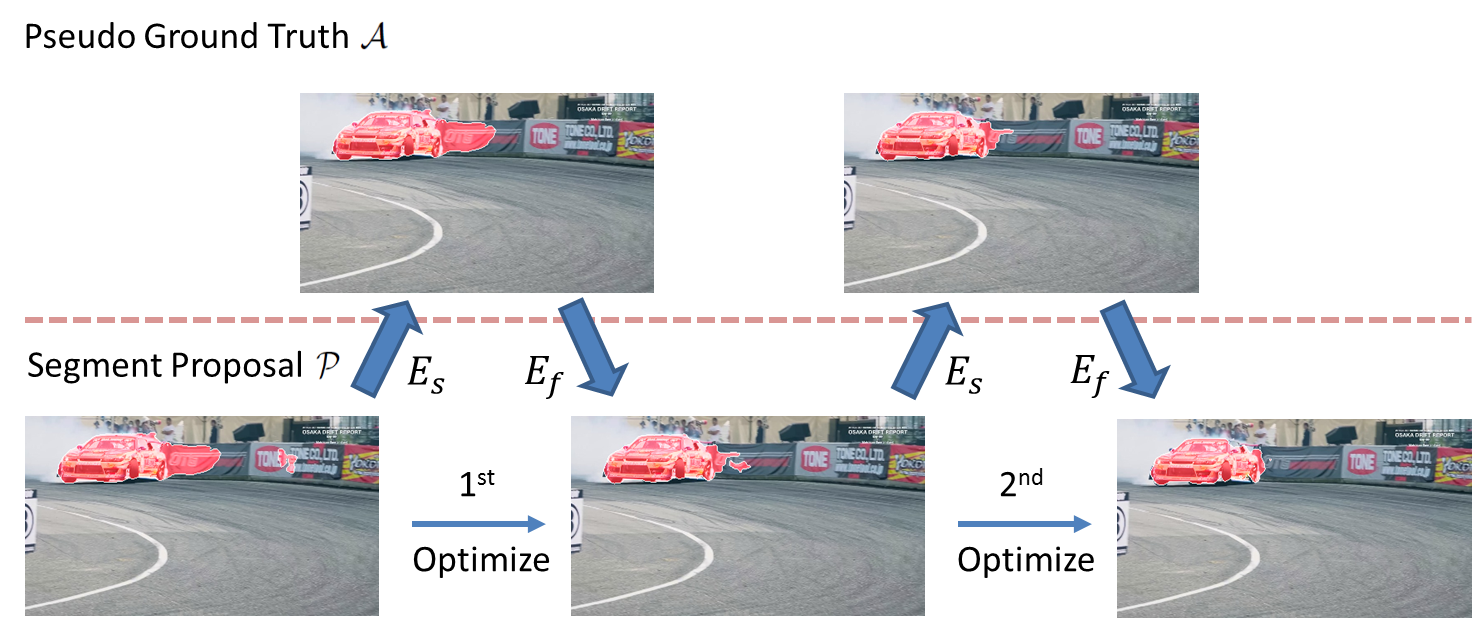}
	\caption{Sample results of iteratively optimizing $E_s$ and $E_f$.
		Starting from an initial set of proposals $\mathcal{P}$, we solve $E_s$ to obtain object-like segments $\mathcal{A}$ as our pseudo ground truths to optimize $E_f$.
		By iteratively updating both energy functions, our algorithm gradually improves the quality of $\mathcal{P}$ and $\mathcal{A}$ to obtain the final segmentation results.
	}
	\label{fig:optimize}
\end{figure}
{\flushleft {\bf Fix $E_f$ and Optimize $E_s$.}}
We first fix the network parameters $\theta$ and optimize $\mathcal{A}$ in $E_s$ of~\eqref{eq:final}.
We adopt a greedy algorithm similar to \cite{Tsai_ECCV_2016}.
Starting from an empty set of $\mathcal{A}$, we add an initial element $a \in \mathcal{V} \backslash \mathcal{A}$ to $\mathcal{A}$ that gives the largest energy gain.
The process is then repeated and stops when one of the following conditions is satisfied:
1) the number of selected proposals reaches a threshold, i.e., $|\mathcal{A}| > N_\mathcal{A}$, and 2) the ratio of the energy gain between two rounds is below a threshold, i.e., $\mathcal{D}(\mathcal{A}^i) < \beta \cdot \mathcal{D}(\mathcal{A}^{i-1})$,
where $\mathcal{D}(\mathcal{A}^i)$ stands for the energy gain, i.e., difference of $E_s$ between two rounds during the optimization process, and $\beta$ is the ratio.
%
{\flushleft {\bf Fix $E_s$ and Optimize $E_f$.}}
Once obtaining $\mathcal{A}$ as the pseudo ground truths, we fix $\mathcal{A}$ and optimize the network with the transferable layer, i.e., $\theta_g$ and $\theta_\mathcal{T}$, in $E_f$ of~\eqref{eq:final}.
We alter the problem to a task that minimizes the network loss $\mathcal{L}$ in an end-to-end fashion, jointly for $\theta_g$ and $\theta_\mathcal{T}$ using the SGD method.
%
{\flushleft {\bf Iterative Optimization.}}
To obtain the final $\mathcal{A}$, $\theta_g$ and $\theta_\mathcal{T}$, instead of directly solving~\eqref{eq:final} for optimal solutions, we solve it via an iterative updating scheme between $E_s$ and $E_f$ until convergence.
%
%
In practice, we measure the intersection-over-union (IoU) of selected segment proposals between two iterations.
%
The optimization process ends when the IoU is larger than a threshold (e.g., 90\%), showing that the set of $\mathcal{A}$ becomes stable.
Fig.~\ref{fig:optimize} shows one example of gradually improved $\mathcal{P}$ and $\mathcal{A}$ via iteratively updating $E_s$ and $E_f$.
The overall process is summarized in Algorithm~\ref{algo:joint}.
\begin{algorithm}[!ht]
	\caption{Unseen Object Segmentation}
	\begin{algorithmic}
		\STATE \textbf{Source Image}: $\mathcal{I}_s$, $Y_s$
		\STATE \textbf{Target Video}: $\mathcal{I}_t$
		
		\STATE \textbf{Initialization}: pre-trained $G_s$ on source inputs, $G_t \leftarrow G_s$, $w_c^{init}$ via~\eqref{eq:sim}
		
		$(\mathcal{P}, \mathcal{F}) \leftarrow \mathcal{T}(G_t(\mathcal{I}_t))$
		
		\WHILE {$\mathcal{P}$ not converged}
		\STATE $\mathcal{A}^0 \leftarrow \emptyset$, $i \leftarrow 1$
		
		\LOOP
		\STATE $a^* = \underset{\{\mathcal{A}^i \in \mathcal{V}\}}{\text{arg max}}$ $E_s(\mathcal{P}, \mathcal{F}; \mathcal{A}^i)$, \text{where} $\mathcal{A}^i \leftarrow \mathcal{A}^{i-1} \cup a$, $a \in \mathcal{V} \backslash \mathcal{A}$
		
		\IF {$|\mathcal{A}| > N_\mathcal{A}$ or $\mathcal{D}(\mathcal{A}^i) < \beta \cdot \mathcal{D}(\mathcal{A}^{i-1})$ \text{when} $i \geq 2$}
		\STATE break
		\ENDIF
		
		\STATE $\mathcal{A}^i \leftarrow \mathcal{A}^{i-1} \cup a^*$, $i \leftarrow i+1$
		\ENDLOOP
		
		\STATE $\mathcal{A} \leftarrow \mathcal{A}^i$
		
		\STATE \text{Optimize} $E_f$: $(\theta_g, \theta_\mathcal{T}) \leftarrow \text{min}$ $\mathcal{L} ( \mathcal{T}(G_t(\mathcal{I}_t)), \mathcal{A} )$
		\STATE $(\mathcal{P}, \mathcal{F}) \leftarrow \mathcal{T}(G_t(\mathcal{I}_t))$
		
		\ENDWHILE
		
		\STATE \textbf{Output}: object segmentation $\mathcal{P}$ of $\mathcal{I}_t$
	\end{algorithmic}
	\label{algo:joint}
\end{algorithm}
%
\section{Experimental Results}
In this section, we first present implementation details of the proposed method, and then we show experimental results on numerous benchmark datasets.
In addition, ablation studies for various components in the algorithm are conducted.
The source code and trained models will be made available to the public.
More results are presented in the supplementary material.
%
\subsection{Implementation Details}
In the submodular function for segment mining, we set $\alpha = 1$ for the facility location term in~\eqref{eq:sub_facility}, and $\lambda_o = 20$, $\lambda_m = 35$ for the unary term in~\eqref{eq:sub_unary}.
During the submodular optimization in~\eqref{eq:sub_combine}, we use $N_\mathcal{A} = 0.8 \cdot|\mathcal{P}|$ and $\beta = 0.8$.
All the parameters are fixed in all the experiments.
For training the CNN model in~\eqref{eq:cnn}, we employ various fully convolutional networks (FCNs)~\cite{Long_CVPR_2015} including the VGG-16~\cite{Simonyan_CoRR_2015} and ResNet-101~\cite{He_CVPR_2016} architectures for both the source and target networks using the Caffe library.
The learning rate, momentum and batch size are set as $10^{-14}$, $0.99$, and $1$, respectively.
%
To further refine segmentation results, we average the responses from the CNN output and a motion prior that is already 
computed in the motion term of~\eqref{eq:sub_unary} to account for both the appearance and temporal information. 
%
\subsection{DAVIS Dataset}
We first conduct experiments on the DAVIS 2016 benchmark dataset~\cite{DAVIS2016}.
Since our goal is to transfer the knowledge from seen categories in images to unseen objects in the video, we manually select all the videos with object categories that are different from the 20 categories in the PASCAL VOC dataset.
In the following, we first conduct ablation studies and experiments to validate the proposed method.
Second, we show that our algorithm can be applied under various settings on the entire set of the DAVIS 2016 dataset.
%
%
{\flushleft {\bf Impact of the Motion Term.}}
One critical component of our framework is to mine useful segments for the further CNN model training step.
In the submodular function of~\eqref{eq:sub_unary}, we incorporate a motion term that accounts for object movements in the video.
To validate its effectiveness, we fix the weight $\lambda_o = 20$ for the appearance and vary the weight $\lambda_m$ for the motion term.
In Table~\ref{tab:selected_iou}, we show the IoU of the selected segment proposals via solving~\eqref{eq:sub_combine} under various values of $\lambda_m$.
The results show that the IoU is gradually improved when increasing the motion weight, which indicates that the quality of selected segments becomes better, and hence we use $\lambda_m = 35$ in all the following experiments.
\begin{table}[t]
	\centering
	\renewcommand{\arraystretch}{0.8}
	\setlength{\tabcolsep}{6pt}
	\caption{IoU of the selected segments with different weights of the motion term on the DAVIS dataset.}
	\label{tab:selected_iou}
	\begin{tabular}{ccccccc}
		\toprule
		$\lambda_m$ & 0 & 5 & 15 & 25 & 35 & 45 \\
		\midrule
		Avg. IoU & 57.2 & 57.4 & 60.5 & 60.6 & 61.0 & 60.3 \\
		\bottomrule
	\end{tabular}
\end{table}
%
%
\begin{table}[t]
	\centering
	\renewcommand{\arraystretch}{1.0}
	\setlength{\tabcolsep}{3.5pt}
	\caption{Results on the DAVIS 2016 dataset with categories excluded from the PASCAL VOC dataset.}
	\label{tab:davis_exclude}
	\begin{tabular}{ccccccccc}
		\toprule
		Methods & bear & bswan & camel & eleph & goat & malw & rhino & Avg. \\
		\midrule
		CVOS~\cite{Taylor_CVPR_2015} & 86.4 & 42.2 & 85.0 & 49.4 & 7.4  & 24.5 & 52.0 & 49.6 \\ 
		MSG~\cite{Ochs_ICCV_2011}  & 85.1 & 52.6 & 75.6 & 68.9 & 73.5 & 4.5  & 90.2 & 64.3 \\ 
		FST~\cite{Papazoglou_ICCV_2013}  & 89.8 & 73.2 & 56.2 & 82.4 & 55.4 & 8.7  & 77.6 & 63.3 \\ 
		NLC~\cite{Faktor_BMVC_2014}  & 90.7 & 87.5 & 76.8 & 51.8 & 1.0  & 76.1 & 68.2 & 64.6 \\ 
		LMP~\cite{Tokmakov_CVPR_2017}  & 69.8 & 50.9 & 78.3 & 78.9 & 75.1 & 38.5 & 76.8 & 66.9 \\ 
		\midrule
		TransferNet~\cite{Hong_CVPR_2016} & 73.7 & \textbf{83.4} & 65.5 & \textbf{76.1} & 78.1 & 17.9 & 42.4 & 62.4 \\ 
		Ours (GloVe) & 82.6 & 67.2 & 68.8 & 61.2 & 70.4 & \textbf{64.7} & 32.0 & 63.8 \\ 
		Ours (init) & 80.3 & 75.6 & 70.9 & 70.4 & 83.1 & 40.9 & 57.7 & 68.4 \\ 
		Ours (opt) & 88.8 & 80.6 & 68.6 & 71.8 & 82.4 & 43.8 & 67.3 & 71.9 \\ 
		Ours (final) & \textbf{89.8} & 76.7 & \textbf{72.0} & 73.8 & \textbf{83.3} & 41.6 & \textbf{71.0} & \textbf{72.6} \\ 
		\midrule
		ARP~\cite{Koh_CVPR_2017} & \textbf{92} & 88.1 & \textbf{90.3} & 84.2 & 77.6 & 58.3 & \textbf{88.4} & 82.7 \\
		FSEG~\cite{Jain_CVPR_2017} & 91.5 & 89.5 & 76.4 & \textbf{86.2} & 84.1 & 83.3 & 77.6 & 84.1 \\
		Ours (ResNet) & 91.8 & \textbf{90.3} & 77.5 & 85.7 & \textbf{84.8} & \textbf{84.9} & 86.0 & \textbf{85.9} \\
		\bottomrule
	\end{tabular}
\end{table}
{\flushleft {\bf Ablation Study.}}
In the middle group of Table~\ref{tab:davis_exclude}, we show the final segmentation results of our method using VGG-16 architecture with various baselines and settings.
We first present a baseline method that uses the GloVe embeddings~\cite{Pennington_EMNLP_2014} to initialize weights, i.e., the similarity between two categories, of the transferable layer.
Since the GloVe is not learned in the image domain between categories, the initialized weights may not reflect the true relationships between the seen and unseen categories, and hence the results are worse than the proposed method for initializing the transferable layer.

\begin{figure}[t]
	\centering
	\begin{tabular}
		{ @{\hspace{0mm}}c@{\hspace{1mm}} @{\hspace{0mm}}c@{\hspace{1mm}} @{\hspace{0mm}}c@{\hspace{1mm}} @{\hspace{0mm}}c@{\hspace{0mm}} }
		
		\includegraphics[width=0.22\linewidth]{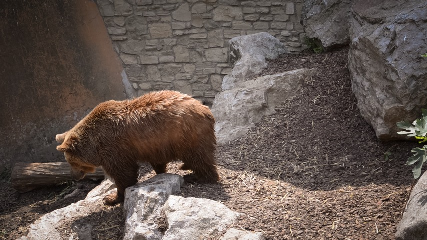} &
		\includegraphics[width=0.22\linewidth]{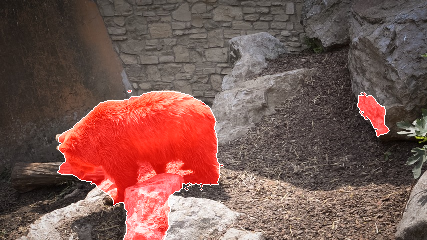} &
		\includegraphics[width=0.22\linewidth]{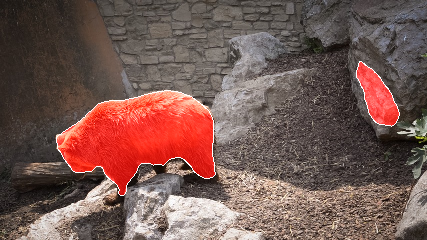} &
		\includegraphics[width=0.22\linewidth]{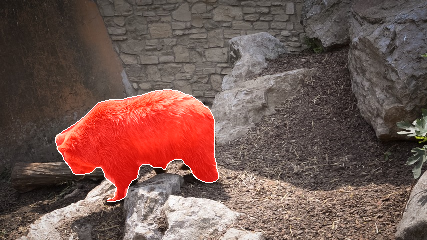} \\

		\includegraphics[width=0.22\linewidth]{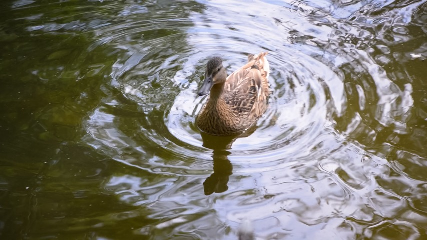} &
		\includegraphics[width=0.22\linewidth]{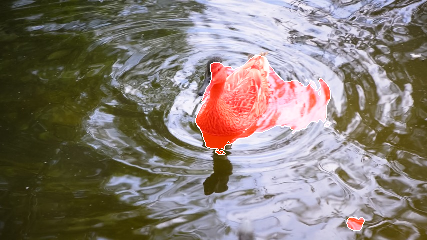} &
		\includegraphics[width=0.22\linewidth]{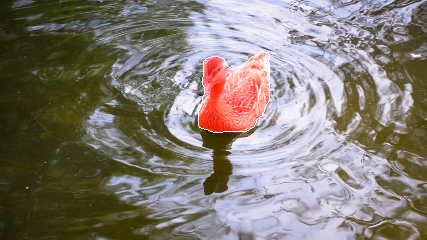} &
		\includegraphics[width=0.22\linewidth]{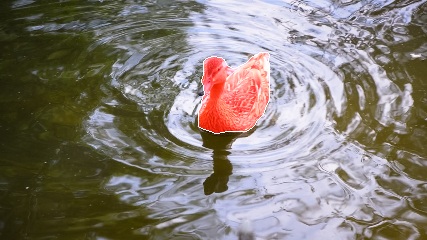} \\

		\includegraphics[width=0.22\linewidth]{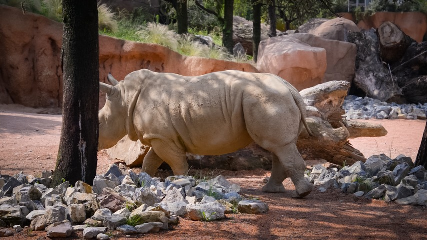} &
		\includegraphics[width=0.22\linewidth]{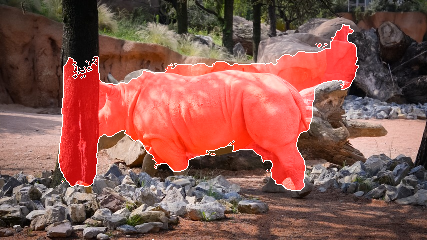} &
		\includegraphics[width=0.22\linewidth]{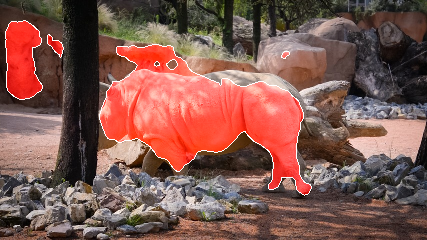} &
		\includegraphics[width=0.22\linewidth]{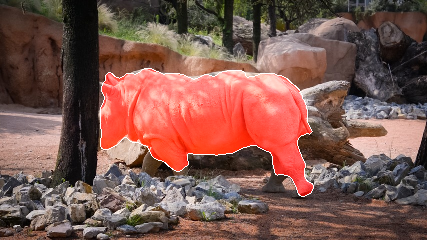} \\

		Input & TransferNet~\cite{Hong_CVPR_2016} & Ours (initial) & Ours (final) \\
		
	\end{tabular}
	\caption{Sample results on the DAVIS dataset with categories excluded from the PASCAL VOC dataset.
		We show that our final results are more accurate in details than the TransferNet~\cite{Hong_CVPR_2016} and with the noisy segments removed from the initial results.
	}
	\label{fig:davis_exclude}
\end{figure}

Furthermore, we show results at different stages, including using the model with initialization before optimizing~\eqref{eq:final}, after optimization, and the final result with motion refinement.
After the optimization, the IoU is improved in 5 out of 7 videos, which shows the effectiveness of the proposed self-learning scheme without using any annotations in the target video.
%
%
{\flushleft {\bf Overall Comparisons.}}
%
%
In Table~\ref{tab:davis_exclude}, we show the comparisons between our method and the state-of-the-art approaches.
We first demonstrate the performance of our method using VGG-16 architecture.
The work closest in the scope to the proposed framework is the method~\cite{Hong_CVPR_2016} that transfers the knowledge between two image domains with mutually exclusive categories in a weakly-supervised setting.
To compare with this approach, we use the authors' public implementation and train the models with the same setting as our method.
We show that our algorithm achieves better IoUs in 5 out of 7 videos and improves the overall IoU by $10.2\%$ on average.
We also note that our model with initialization already performs favorably against \cite{Hong_CVPR_2016}, which demonstrates that the proposed transferable layer is effective in learning knowledge from seen categories to unseen ones.
Visual comparisons are presented in Fig.~\ref{fig:davis_exclude}.

In addition, we present more results of video object segmentation methods in Table~\ref{tab:davis_exclude} and show that the proposed algorithm achieves better performance.
Different from existing approaches that rely on long-term trajectory~\cite{Taylor_CVPR_2015,Ochs_ICCV_2011} or motion saliency~\cite{Papazoglou_ICCV_2013,Faktor_BMVC_2014} to localize foreground objects, we use the proposed self-learning framework to segment unseen object categories via transfer learning.
We note that the proposed method performs better than the CNN-based model~\cite{Tokmakov_CVPR_2017} that utilizes synthetic videos with pixel-wise segmentation annotations.

We further employ the stronger ResNet-101 architecture and compare with state-of-the-art unsupervised video object segmentation methods.
In the bottom group of Table~\ref{tab:davis_exclude}, we show that our approach performs better than FSEG~\cite{Jain_CVPR_2017} using the same architecture and training data from PASCAL VOC, i.e., the setting of the appearance stream in FSEG~\cite{Jain_CVPR_2017}.
In addition, compared to ARP~\cite{Koh_CVPR_2017} that adopts a non-learning based framework via proposal post-processing and is specifically designed for video object segmentation, our algorithm performs better and is flexible under various settings such as using weakly-supervised signals.
%
{\flushleft {\bf Results on the Entire DAVIS 2016 Dataset.}}
In addition to performing object segmentation on unseen object categories, our method can adapt to the weakly-supervised setting by simply initializing the weights in the transferable layer as a one-hot vector, where only the known category is set to 1 and the others are 0.
We evaluate this setting on the DAVIS 2016 dataset with categories shared in the PASCAL VOC dataset. Note that, we still adopt the unsupervised setting for the unseen categories. The results on the entire DAVIS 2016 dataset are shown in Table~\ref{tab:davis_2016}.
In comparison with a recent weakly-supervised method~\cite{Zhang_CVPR_2017} and the baseline model~\cite{Long_CVPR_2015} (our initial result), our approach addresses the transfer learning problem and outperforms their methods by $6.5\%$ and $6.1\%$, respectively.

\begin{figure}[t]
	\centering
	\begin{tabular}
		{ @{\hspace{0mm}}c@{\hspace{1mm}} @{\hspace{0mm}}c@{\hspace{1mm}} @{\hspace{0mm}}c@{\hspace{1mm}} @{\hspace{0mm}}c@{\hspace{0mm}} }
		
		\includegraphics[width=0.22\linewidth]{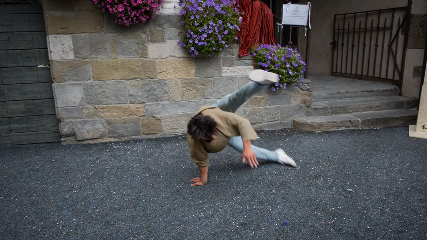} &
		\includegraphics[width=0.22\linewidth]{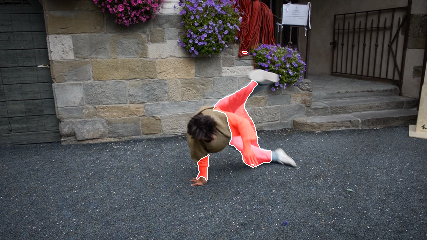} &
		\includegraphics[width=0.22\linewidth]{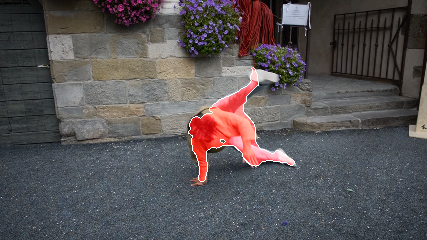} &
		\includegraphics[width=0.22\linewidth]{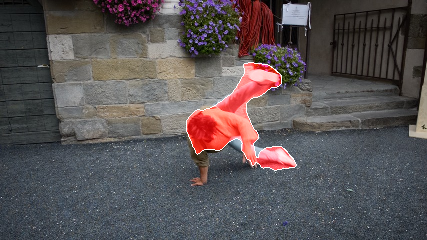} \\
		
		\includegraphics[width=0.22\linewidth]{figure/davis_same/cows_00049.png} &
		\includegraphics[width=0.22\linewidth]{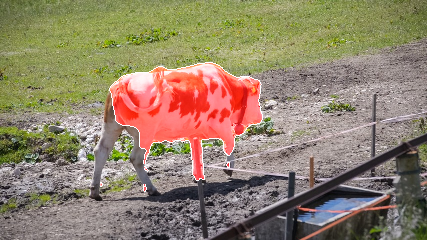} &
		\includegraphics[width=0.22\linewidth]{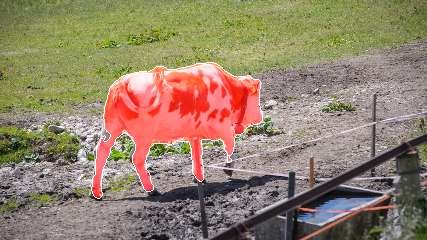} &
		\includegraphics[width=0.22\linewidth]{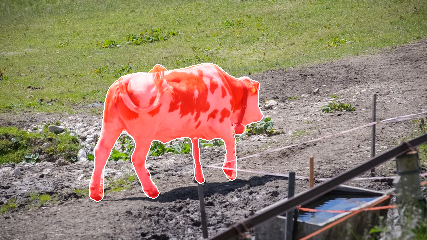} \\
		
		\includegraphics[width=0.22\linewidth]{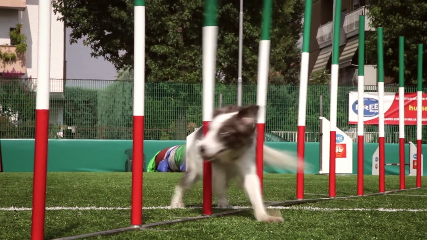} &
		\includegraphics[width=0.22\linewidth]{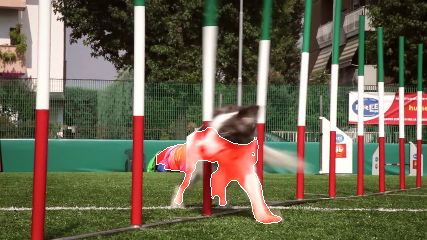} &
		\includegraphics[width=0.22\linewidth]{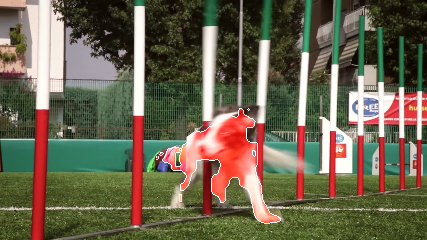} &
		\includegraphics[width=0.22\linewidth]{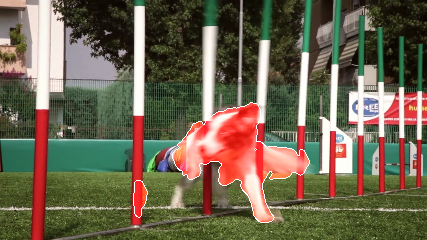} \\

		Input & FCN~\cite{Long_CVPR_2015} & Ours & Ours (no sup.) \\
		
	\end{tabular}
	\caption{Segmentation results on the DAVIS dataset with categories shared in the PASCAL VOC dataset.
		We show that both of our results with and without supervision have more complete object segmentations than the baseline FCN model~\cite{Long_CVPR_2015} (our initial result) that uses the weak supervision.
	}
	\label{fig:davis_include}
\end{figure}

Although the same categories are shared between the source and target domains in this setting, we can still assume that the object category is unknown in the target video.
Under this fully unsupervised setting without using any pixel-wise annotations in videos during training, we show that our method improves the results of FSEG~\cite{Jain_CVPR_2017} and other unsupervised algorithms~\cite{Papazoglou_ICCV_2013,Faktor_BMVC_2014}.
Sample results are presented in Fig.~\ref{fig:davis_include}.
%
%
\begin{table}[!hb]
	\centering
	\renewcommand{\arraystretch}{1}
	\setlength{\tabcolsep}{3.5pt}
	\caption{Results on the entire DAVIS 2016 dataset.}
	\label{tab:davis_2016}
	\begin{tabular}{cccccccc}
		\toprule
		& \multicolumn{3}{c}{Weak Supervision} & \multicolumn{4}{c}{No Supervision} \\
		\cmidrule(lr){2-4} \cmidrule(lr){5-8}
		Methods & SPFTN~\cite{Zhang_CVPR_2017} & FCN \cite{Long_CVPR_2015} & Ours & FST~\cite{Papazoglou_ICCV_2013} & NLC~\cite{Faktor_BMVC_2014} & FSEG~\cite{Jain_CVPR_2017} & Ours \\
		\midrule
		Avg. IoU & 61.2 & 61.6 & 67.7 & 57.5 & 64.1 & 64.7 & 66.5 \\
		\bottomrule
	\end{tabular}
\end{table}
\subsection{YouTube-Objects Dataset}
We evaluate our method on the YouTube-Objects dataset~\cite{Prest_CVPR_2012} with annotations provided by \cite{Jain_ECCV_2014} for 126 videos.
Since this dataset contains 10 object categories that are shared with the PASCAL VOC dataset, we conduct experiments using the weakly-supervised setting.
In Table~\ref{tab:youtube}, we compare our method with the state-of-the-art algorithms that use the class-level weak supervision.
With the VGG-16 architecture, the proposed framework performs well in 6 out of 10 categories and achieves the best IoU on average.
Compared to the baseline FCN model~\cite{Long_CVPR_2015} used in our algorithm, there is a performance gain of $9\%$.
In addition, while existing methods rely on training the segment classifier~\cite{Tang_CVPR_2013}, integrating object proposals with detectors~\cite{Zhang_CVPR_2015}, co-segmentation via modeling relationships between videos~\cite{Tsai_ECCV_2016}, or self-paced fine-tuning~\cite{Zhang_CVPR_2017}, the proposed method utilizes a self-learning scheme to achieve better segmentation results.
With the ResNet-101 architecture, we compare our method with DeepLab~\cite{CP2016Deeplab} and FSEG~\cite{Jain_CVPR_2017}. We show that the proposed method improves the performance in 6 out of 10 categories and achieves the best averaged IoU.
%
\begin{table}[t]
	\centering
	\renewcommand{\arraystretch}{1}
	\setlength{\tabcolsep}{2.5pt}
	\caption{Results on the YouTube-Objects dataset.}
	\label{tab:youtube}
	\begin{tabular}{cccccccccccc}
		\toprule
		Methods & aero & bird & boat & car & cat & cow & dog & horse & mbike & train & Avg. \\
		\midrule
		DSA~\cite{Tang_CVPR_2013} & 17.8 & 19.8 & 22.5 & 38.3 & 23.6 & 26.8 & 23.7 & 14.0 & 12.5 & 40.4 & 23.9 \\
		FCN~\cite{Long_CVPR_2015} & 68.3 & 65.7 & 55.7 & 76.6 & 52.3 & 50.4 & 55.6 & 52.6 & 35.7 & 55.9 & 56.9 \\
		DET~\cite{Zhang_CVPR_2015} & 72.4 & 66.6 & 43.0 & 58.9 & 36.4 & 58.2 & 48.7 & 49.6 & 41.4 & 49.3 & 52.4 \\
		CoSeg~\cite{Tsai_ECCV_2016} & 69.3 & \textbf{76.1} & 57.2 & 70.4 & \textbf{67.7} & 59.7 & 64.2 & 57.1 & 44.1 & 57.9 & 62.3 \\
		SPFTN~\cite{Zhang_CVPR_2017} & \textbf{81.1} & 68.8 & 63.4 & 73.8 & 59.7 & 64.5 & 63.4 & 58.2 & \textbf{52.4} & 45.5 & 63.1 \\
		Ours (VGG) & 74.6 & 65.3 & \textbf{66.9} & \textbf{79.5} & 64.2 & \textbf{68.3} & \textbf{67.3} & \textbf{61.7} & 51.5 & \textbf{59.4} & \textbf{65.9} \\
		\midrule
		DeepLab~\cite{CP2016Deeplab} & 80.6 & 67.8 & 66.9 & 73.3 & 55.3 & 61.8 & 63.9 & 45.5 & 54.7 & 56.4 & 62.6 \\
		FSEG~\cite{Jain_CVPR_2017} & 83.4 & 60.9 & \textbf{72.6} & 74.5 & \textbf{68.0} & 69.6 & 69.1 & \textbf{62.8} & 61.9 & \textbf{62.8} & 68.6 \\
		Ours (ResNet) & \textbf{83.5} & \textbf{76.4} & 70.0 & \textbf{75.3} & 65.9 & \textbf{69.7} & \textbf{71.6} & 54.7 & \textbf{63.8} & 58.7 & \textbf{69.0} \\
		\bottomrule
	\end{tabular}
\end{table}
%
\section{Concluding Remarks}
%
In this paper, we propose a self-learning framework to segment objects in unlabeled videos.
By utilizing existing annotations in images, we design a model to adapt seen object categories from source images to the target video.
The entire process is decomposed into two sub-problems: 1) a segment mining module to select object-like proposals, and 2) a CNN model with a transferable layer that adapts feature representations for target videos.
To optimize the proposed formulation, we adopt an iterative scheme to obtain final solutions.
Extensive experiments and ablation study show the effectiveness of the proposed algorithm against other state-of-the-art methods on numerous benchmark datasets.
{\flushleft {\bf Acknowledgments.}}
This work is supported in part by Ministry of Science and Technology under grants MOST 105-2221-E-001-030-MY2 and MOST 107-2628-E-001-005-MY3.



\bibliographystyle{splncs04}
\bibliography{mybib}

\end{document}